\documentclass[11pt]{article}
\usepackage{coling2018}
\usepackage{times}
\usepackage{url}
\usepackage{latexsym}
\usepackage{amsmath, amsfonts, amssymb}
\usepackage{booktabs, multirow, tabularx, makecell}
\usepackage{siunitx}
\usepackage{graphicx}

\newcommand{\R}{\mathbb{R}}

\title{Extracting Parallel Sentences with Bidirectional Recurrent Neural Networks to Improve Machine Translation}

\author{Francis Gr\'egoire \\
  Universit\'e de Montr\'eal \\
  {\tt francis.gregoire@rd.mila.quebec} \\\And
  Philippe Langlais \\
  Universit\'e de Montr\'eal \\
  {\tt felipe@iro.umontreal.ca} \\}

\date{}

\begin{document}
\maketitle
\begin{abstract}
Parallel sentence extraction is a task addressing the data sparsity problem found in multilingual natural language processing applications. We propose a bidirectional recurrent neural network based approach to extract parallel sentences from collections of multilingual texts. Our experiments with noisy parallel corpora show that we can achieve promising results against a competitive baseline by removing the need of specific feature engineering or additional external resources. To justify the utility of our approach, we extract sentence pairs from Wikipedia articles to train machine translation systems and show significant improvements in translation performance.
\end{abstract}

\section{Introduction}
\label{sect:introduction}
\blfootnote{
    \hspace{-0.65cm}  
    This work is licensed under a Creative Commons 
    Attribution 4.0 International License.
    License details:
    \url{http://creativecommons.org/licenses/by/4.0/}
}
Parallel corpora are a prerequisite for many multilingual natural language processing (NLP) applications. Unfortunately, parallel data is available for a relatively small number of language pairs and for few specific domains. With the increasing amount of comparable corpora on the World Wide Web, a potential solution to alleviate the parallel data sparsity issue is to extract parallel sentences from this more abundant source of information. Therefore, the objective of parallel sentence extraction is to extract parallel sentences from such comparable corpora to increase the quantity of parallel data and the range of the covered domains. Comparable corpora can be defined as collections of topic-aligned but non-sentence-aligned multilingual texts.

Recent advances in deep learning architectures with recurrent neural networks (RNNs) have shown that they can successfully learn complex mappings from variable-length sequences to continuous vector representations. While numerous natural language processing tasks have successfully applied those models, ranging from handwriting generation~\cite{Graves:2013} to machine comprehension~\cite{Hermann:2015}, most of the multilingual efforts have been devoted to machine translation~\cite{Sutskever:2014,Cho:2014}, although more research interests have been recently devoted to multilingual semantic textual similarity.\footnote{http://alt.qcri.org/semeval2017/task2/}

In this paper, we propose a parallel sentence extraction system to measure the translational equivalence between sentences in two languages. Our system is based on bidirectional recurrent neural networks that can learn sentence representations in a shared vector space by explicitly maximizing the similarity between parallel sentences. In contrast to previous approaches, by leveraging these continuous vector representation of sentences we remove the need to rely on multiple models and specific feature engineering. Experiments on noisy parallel corpora show that our approach outperforms a competitive baseline. To justify the utility of our approach, we add the sentence pairs extracted from Wikipedia articles to a parallel corpus to train machine translation systems and show improvements in translation performance. Our experimental results lead us to believe that our system is a promising tool to create new aligned multilingual resources.

\section{Related work}
\label{sect:related-work}
Traditional systems developed to extract parallel sentences from comparable corpora typically rely on multiples models or metadata from articles structure. Munteanu and Marcu~\shortcite{Munteanu:2005} present a complete system based on statistical word alignment models and a maximum entropy classifier to automatically extract sentence pairs from newspaper collections. The authors evaluate the quality of the extracted sentences by showing that they improve the performance of statistical machine translation (SMT) systems. The overall structure of this approach is still considered state-of-the-art. Adafre and Rijke~\shortcite{Adafre:2006} is the first work to observe that articles from Wikipedia are likely to generate parallel corpora useful for machine translation. These two approaches are extended by Smith et al.~\shortcite{Smith:2010} where the authors introduce several new features by exploiting the structure and metadata of Wikipedia article pairs. They use their augmented set of features in a conditional random field and obtain state-of-the-art results on a small set of 20 manually annotated Wikipedia article pairs. The work of Abdul-Rauf and Schwenk~\shortcite{Rauf:2009} proposes a different approach, in which they use an SMT system built from a small parallel corpus. Instead of using a classifier, they translate the source language side of a comparable corpus to find candidate sentences on the target language side. They determine if a translated source sentence and a candidate target sentence are parallel by measuring the word error rate and the translation error rate. A simplification of these systems is proposed in~\cite{Azpeitia:2017}, where the similarity between two sentences is defined as the average of the Jaccard similarity coefficients obtained between sentence token sets and lexical translations determined by IBM models~\cite{Brown:1993}. Their approach also combine various features, such as longest common prefix matching, numbers and capitalized truecased tokens.

Chu et al.~\shortcite{Chu:2016} train a neural machine translation (NMT) system to generate sentence representations with the encoder, which are then used as additional features to the system of Munteanu and Marcu~\shortcite{Munteanu:2005}. Similarly, Cristina et al.~\shortcite{Cristina:2017} study sentence representations obtained from the encoder of an NMT system to detect new parallel sentence pairs. By comparing cosine similarities, they show that they can distinguish parallel and non-parallel sentences. A different approach exploiting continuous vector representations is proposed in Grover and Mitra~\shortcite{Grover:2017}. After learning word representations using the bilingual word embeddings model of~\cite{Luong:2015}, they use a convolutional neural network on a similarity matrix to classify if a pair of sentences is aligned or not. These approaches are different from ours, where we use a single end-to-end model to estimate the conditional probability distribution that two sentences are parallel.

\section{Approach}
\label{sect:approach}
\subsection{Negative sampling}
\label{ssec:negative-sampling}
As positive examples, we use a parallel corpus \(C\) consisting of \(n\) parallel sentence pairs \(\left\lbrace(\mathbf{s}^{S}_{k}, \mathbf{s}^{T}_{k})\right\rbrace_{k=1}^{n}\), where \(S\) and \(T\) denote the source sentences and target sentences. Since we want a model that learns differentiable vector representations to distinguish parallel from non-parallel sentences, we use negative sampling to generate negative examples. Therefore, we randomly sample \(m\) non-parallel target sentences for every positive source sentence, such that \((\mathbf{s}^{S}_{k}, \mathbf{s}^{T}_{j})\) for \(j \neq k\). This process is repeated at the beginning of each training epoch to allow the model to learn on a larger number of non-parallel sentence pairs. Hence, our training set contains \(n(1 + m)\) triples \((\mathbf{s}^{S}_{i}, \mathbf{s}^{T}_{i}, y_{i})\), where \(\mathbf{s}^{S}_{i} = (w^{S}_{i,1}, \dots, w^{S}_{i,N})\) is a source sentence of \(N\) tokens, \(\mathbf{s}^{T}_{i} = (w^{T}_{i,1}, \dots, w^{T}_{i,M})\) is a target sentence of \(M\) tokens, and \(y_{i}\) is the label representing the translation relationship between \(\mathbf{s}^{S}_{i}\) and \(\mathbf{s}^{T}_{i}\), so that \(y_{i} = 1\) if \((\mathbf{s}^{S}_{i}, \mathbf{s}^{T}_{i}) \in C\) and \(y_{i} = 0\) otherwise. 

The advantage of negative sampling is its simplicity. However, relying only on randomness to select negative sentence pairs makes most of the examples very non-parallel and easy to classify. An interesting way to generate negative examples would be to replace only a segment of a sentence. Similarly, we could replace a sentence from a parallel pair with another sentence that is close to it in the vector space. This would make the problem harder, but potentially could make the classifier stronger. We leave such investigations as future work.

\subsection{Model}
\label{ssec:model}
\begin{figure}[t]
  \centering
  \includegraphics[width=0.60\textwidth]{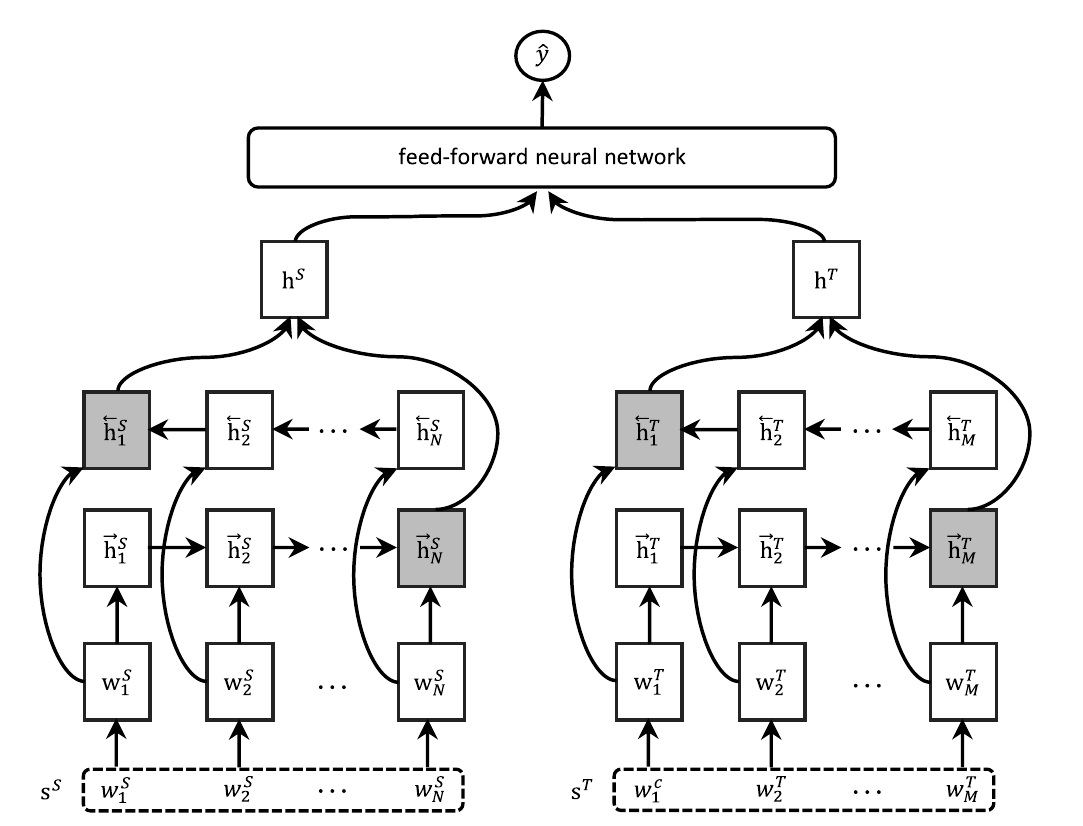}
  \caption{Graphical illustration of the siamese bidirectional recurrent neural networks. The final recurrent state of the forward and backward RNNs are concatenated and then fed into a feed-forward neural network culminating in a sigmoid output layer.}
  \label{fig:model}
\end{figure}

Our idea is to use RNNs to learn cross-language semantics between sentence pairs to estimate the probability that they are translations of each other. The proposed model architecture consists of a bidirectional RNNs (BiRNN)~\cite{Schuster:97} sentence encoder with recurrent activation functions such as long short-term memory units (LSTM)~\cite{Hochreiter:97} or gated recurrent units (GRU)~\cite{Cho:2014}. Since we want vector representations in a shared vector space we use a siamese network~\cite{Bromley:1993} with tied weights, which is equivalent to using a single BiRNN to encode a pair of sentences into two continuous vector representations. The source and target sentence representations are then fed into a feed-forward neural network with a sigmoid output layer which calculates the probability that they are parallel. The architecture of our approach is illustrated in Figure~\ref{fig:model}.

To avoid repetition and for clarity, we only define equations of the BiRNN encoding the source sentence. For the target sentence, simply substitute \(S\) for \(T\). At each time step \(t\), the token in the \(i\)-th sentence, \(w^{S}_{i,t}\), defined by its integer index \(k\) in the vocabulary \(V^{S}\), is represented as a one-hot vector, \(\mathbf{x}_{k}^{S} \in \left\{0,1\right\}^{|V|}\). This one-hot vector is multiplied with an embedding matrix, \(\mathbf{E}^{S} \in \R^{|V^{S}| \times d_{e}}\), to get a continuous vector representation of this token, \(\mathbf{w}^{S}_{i, t} \in \R^{d_{e}}\), which serves as an input for the forward and backward recurrent states of the BiRNN encoder, \(\overrightarrow{\mathbf{h}}^{S}_{i, t}\) and \(\overleftarrow{\mathbf{h}}^{S}_{i, t}\). The forward RNN reads the variable-length sentence and updates its recurrent state from the first token until the last one to create a fixed-size continuous vector representation of the sentence, \(\mathbf{h}^{S}_{i, N} \in \R^{d_{h}}\). The backward RNN processes the sentence in reverse. We use the concatenation of the last recurrent state in both directions as a final representation \(\mathbf{h}^{S}_{i}\ = [\overrightarrow{\mathbf{h}}^{S}_{i,N}\ ; \overleftarrow{\mathbf{h}}^{S}_{i,1}]\).\footnote{We considered combining the recurrent states with mean pooling and max pooling to obtain a fixed-size vector representation, but obtained inferior performance.} The steps we described to encode a sentence are defined as:
\begin{gather}
  \mathbf{w}^{S}_{i,t} = \mathbf{E}^{S^\top} \mathbf{w}^{S}_{k}, \\
  \overrightarrow{\mathbf{h}}^{S}_{i,t} = \phi(\overrightarrow{\mathbf{h}}^{S}_{i, t-1},
    \mathbf{w}^{S}_{i,t}), \\
  \overleftarrow{\mathbf{h}}^{S}_{i,t} = \phi(\overleftarrow{\mathbf{h}}^{S}_{i, t+1},
    \mathbf{w}^{S}_{i,t}),
\end{gather}
where \(\phi(\cdot)\) is an LSTM or GRU. 

After both source and target sentences have been encoded, we capture their matching information by using their element-wise product and absolute element-wise difference. We estimate the conditional probability that the sentences are parallel by feeding the matching vectors into a feed-forward neural network with a sigmoid output layer:
\begin{gather}
  \mathbf{h}_{i}^{(1)} = \mathbf{h}^{S}_{i} \odot \mathbf{h}^{T}_{i}, \\
  \mathbf{h}_{i}^{(2)} = |\mathbf{h}^{S}_{i} - \mathbf{h}^{T}_{i}|, \\
  \mathbf{h}_{i} = \tanh(\mathbf{W}^{(1)}\mathbf{h}_{i}^{(1)} + \mathbf{W}^{(2)}\mathbf{h}_{i}^{(2)} + \mathbf{b}), \\
  p(y_{i}=1|\mathbf{h}_{i}) = \sigma(\mathbf{v}\mathbf{h}_{i} + b),
\end{gather}
where \(\sigma(\cdot)\) is the sigmoid function, \(\mathbf{W}^{(1)} \in \R^{d_{f} \times d_{h}}\), \(\mathbf{W}^{(2)} \in \R^{d_{f} \times d_{h}}\), \(\mathbf{v} \in \R^{d_{f}}\), \(\mathbf{b} \in \R^{d_{f}}\) and \(b\) are model parameters. The value \(d_{f}\) is the size of the hidden layer in the feed-forward neural network.

The model is trained by minimizing the cross entropy of our labeled sentence pairs:
\begin{equation}
   \mathcal{L} = -\sum^{n(1+m)}_{i=1} \Big(y_{i} \log \sigma(\mathbf{v}\mathbf{h}_{i} + b) + (1-y_{i}) \log (1-\sigma(\mathbf{v}\mathbf{h}_{i} + b))\Big).
\end{equation}

For prediction, a sentence pair is classified as parallel if its probability is greater than or equal to a decision threshold \(\rho\) that we need to fix:
\begin{equation}
  \hat{y}_{i} =
    \begin{cases}
      1 & \text{if}\ p(y_{i}=1|\mathbf{h}_{i}) \geq \rho, \\
      0 & \text{otherwise}.
    \end{cases}
\end{equation}

The two embedding matrices \(\mathbf{E}^{S}\) and \(\mathbf{E}^{T}\) are parameters of the model that we must learn. As the size of the vocabulary and the dimension \(d_{e}\) increase, the number of parameters can become considerably expensive to estimate. For that reason, it is common to initialize the embedding matrices using word embeddings pre-trained on a large collection of texts.\footnote{We pre-trained bilingual word embeddings as proposed by~\cite{Gouws:2015} and~\cite{Smith:2017}. While we observed faster learning and a small improvement in the performance of our model, we prefer to learn the embedding matrices from scratch instead of relying on additional external resources.}

\section{Experiments and Results}
\label{sect:experiments}
To assess the effectiveness of our approach, we compare it in multiple settings. In Section~\ref{ssec:model-evaluation}, we measure its capacity to extract parallel sentences found in parallel corpora in which we inserted non-parallel sentences. We extract sentence pairs from real comparable corpora in Section~\ref{ssec:machine-translation} and validate their utility by measuring their impact on SMT and NMT systems.

\subsection{System comparison}
\label{ssec:model-evaluation}
\subsubsection{Data}
\label{sssec:data-1}
To create our training and test sets we use the WMT'15 English--French datasets.\footnote{http://www.statmt.org/wmt15/translation-task.html} The positive examples of our training set consist of 500,000 parallel sentence pairs randomly selected from the Europarl corpus~\cite{Koehn:2005}. To generate negative examples, at the beginning of each training epoch we sample \(m\) non-parallel sentences per positive example, as described in Section~\ref{ssec:negative-sampling}. The vocabulary size is 69,381 for English and 84,182 for French.

The most reliable way to create test sets to compare different approaches would be to have professional translators manually annotate parallel sentences from comparable corpora. However, this option is expensive and impractical. Therefore, it is common practice to compare parallel sentence extraction systems on artificially-created noisy parallel data by inserting non-parallel sentences into a parallel corpus. Thus, to create our test sets we use parallel sentences from the newstest2012 and Europarl corpora and insert artificial noise by substituting target sentences with non-parallel target sentences from their respective held-out dataset. We sample 1,000 parallel sentences from the newstest2012 corpus and create ten test sets with a noise ratio \(r \in \{0\%, 10\%, \dots, 80\%, 90\%\}\), where \(r\) is the ratio of artificial non-parallel sentences. The same process is repeated with the Europarl corpus to create three test sets with \(r \in \{0\%, 50\%, 90\%\}\). Each final test set consists of 1,000,000 sentence pairs generated from the Cartesian product between the sentences in both languages. For example, 400 out of the 1,000 sentence pairs are parallel if \(r=60\%\), in which only 0.04\% of the 1,000,000 sentence pairs generated from the Cartesian product are truly parallel.

We tokenize all datasets with the scripts from Moses~\cite{Moses:2007}.\footnote{https://github.com/moses-smt/mosesdecoder} The maximum sentence length is set to 80 tokens. Each out-of-vocabulary word is mapped to a special UNK token.

\subsubsection{Evaluation metrics}
\label{sssec:eval-metrics-1}
For evaluating the performance of our models, a sentence pair predicted as parallel is correct if it is present in the set of parallel sentences of the test set. Precision is the proportion of truly parallel sentence pairs among all extracted sentence pairs. Recall is the proportion of truly parallel extracted sentence pairs among all parallel sentence pairs in the test set. The F\(_{1}\) score is the harmonic mean of precision and recall.

\subsubsection{Baseline}
\label{sssec:baseline-1}
For comparison, we use a parallel sentence extraction system developed in-house based on the works of~\cite{Munteanu:2005} and~\cite{Smith:2010}. The system consists of a candidate sentence pair filtering process and three models; two word alignment models and a maximum entropy classifier. The word alignment models are trained on both language directions using our training set of 500,000 parallel sentence pairs. To train the classifier, we select another 100,000 parallel sentence pairs from the held-out Europarl dataset. To generate negative examples, we use 100,000 non-parallel sentence pairs that have successfully passed the candidate sentence pair selection process.\footnote{During our experiments, we did not observe any significant gain by using more than 100,000 parallel sentence pairs.}

\noindent\textbf{Candidate sentence pair selection}\hspace{1mm} A sentence pair filtering process is used to select a fixed number of negative sentence pairs to train the maximum entropy classifier. The objective is to select similar non-parallel sentence pairs to make the classifier more robust to noise. It is also used during prediction to filter out the unlikely sentence pairs of the Cartesian product. Since most of these sentence pairs are not parallel, the filtering process significantly reduces the number of sentence pairs to evaluate. The filtering process consists of a two-step procedure. First, it verifies that the ratio of the lengths between two sentences is not greater than 2.\footnote{Most parallel sentence extraction systems use a length ratio value equals to 2.} It then uses a word-overlap filter to check for both sentences that at least 50\% of their words have a translation in the other sentence. Every pair that does not fulfill these two conditions is discarded. The filtering process is only applied to the baseline model.

\noindent\textbf{Word alignment models}\hspace{1mm} The translation and alignment tables are estimated using the HMM alignment model of~\cite{Vogel:1996}. These probability tables are required to calculate the value of many alignment features used in the classifier.

\noindent\textbf{Maximum entropy classifier}\hspace{1mm} The classifier relies on word-level alignment features between two sentences, such as the number of connected words, top three largest fertilities, length of the longest connected substring, log probability of the alignments, and also general features, such as the lengths of the sentences, length difference and the percentage of words on each side that have a translation on the other side. For each sentence pair, a total of 31 features must be calculated and it is classified as parallel if the classifier outputs a probability score greater than or equal to a decision threshold \(\rho\) which needs to be fixed.

\subsubsection{Training settings}
\label{sssec:training-1}
The models of our approach are implemented in TensorFlow~\cite{Tensorflow:2016}. We use a BiRNN with a single layer in each direction with 512-dimensional word embeddings and 512-dimensional recurrent states. We use LSTM as recurrent activation functions. The hidden layer of the feed-forward neural network has 256 hidden units. To train our models, we use Adam~\cite{Kingma:2014} with a learning rate of 0.0002 and a minibatch of 128 examples. Models are trained for a total of 15 epochs. To avoid exploding gradients, we apply gradient clipping such that the norm of all gradients is no larger than 5~\cite{Pascanu:2013}. We apply dropout to prevent overfitting with a probability of 0.2 and 0.3 for the non-recurrent input and output connections respectively~\cite{Zaremba:2014}.

\subsubsection{Number of negative examples}
\label{sssec:n-negative}
\begin{figure}[t]
  \includegraphics[width=0.45\textwidth]{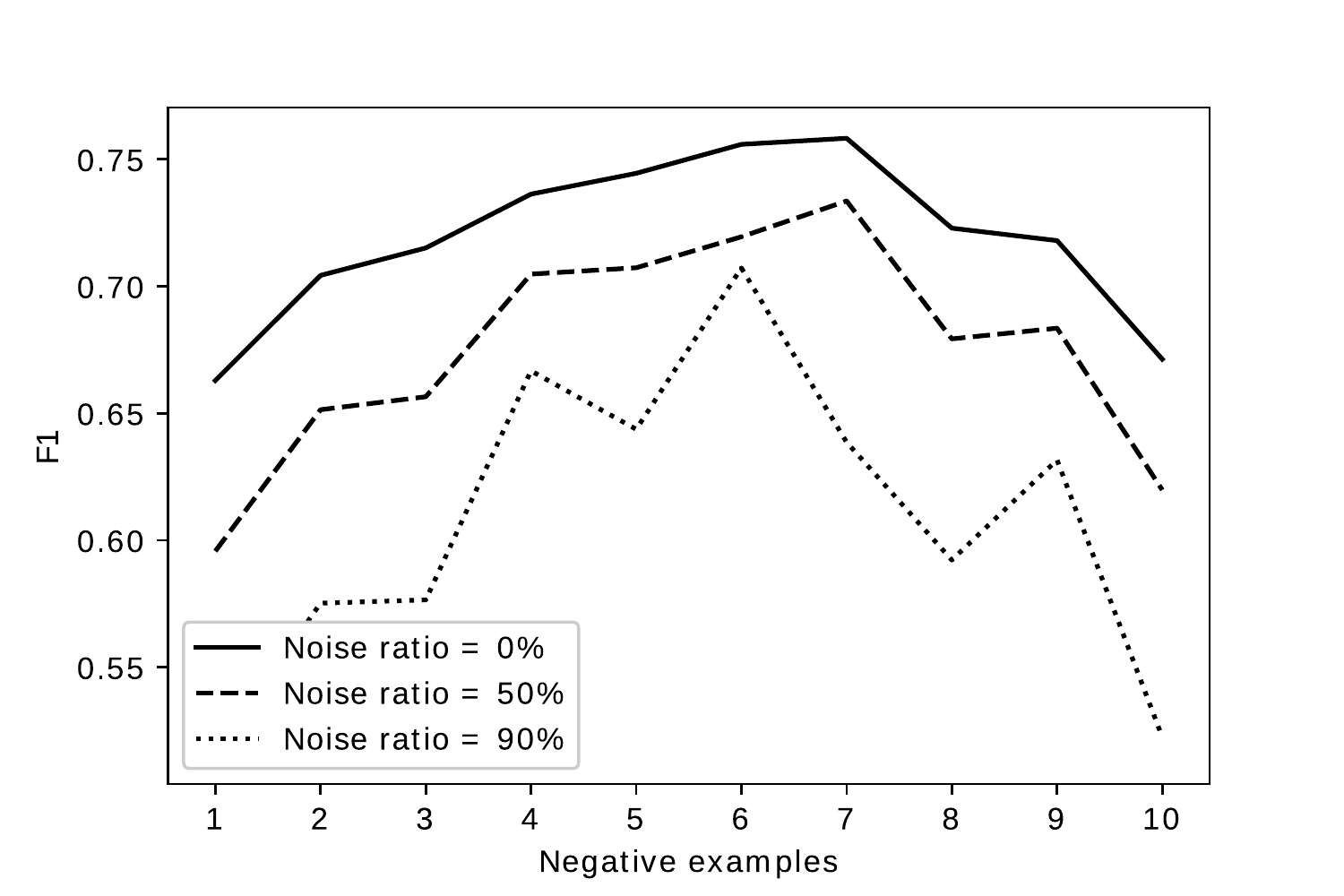}
  \centering
  \caption{F\(_{1}\) scores of BiRNN as we increase the number of negative examples \(m\) per parallel sentence in the training set. The models are evaluated on newstest2012 with noise ratios of 0\%, 50\% and 90\%.}
  \label{fig:n-negatives}
\end{figure}

We evaluate the performance of our approach as we increase the number of negative examples \(m\) per parallel sentence pair in our training set. We trained a total of 10 models with \(m \in \{1,\dots,10\}\), such that with \(m=1\) and \(m=10\) a model is respectively trained on 1,000,000 and 5,500,000 sentence pairs per epoch, with a positive to negative sentence pairs ratio of 50\% and 9\%. The more the value of \(m\) increases, the more the training set becomes unbalanced. We know that extracting parallel sentences from comparable corpora in practice is an unbalanced classification task in which non-parallel sentences represent the majority class. Although an unbalanced training set is not desired since a classifier trained on such data will typically tend to predict the majority class and have a poor precision, the overall impact on the performance of BiRNN is not clear. Figure~\ref{fig:n-negatives} shows the F\(_{1}\) scores of BiRNN with respect to the value of \(m\) evaluated on newstest2012 with noise ratios of 0\%, 50\% and 90\%. Each reported F\(_{1}\) score is the one calculated at the decision threshold value maximizing the area under the precision-recall curve. For \(m\geq7\), we observe a deterioration in the performance of our systems evaluated on test sets with noise ratios of 0\% and 50\%, whereas \(m=6\) is the best performing model on the test set with a noise ratio of 90\%. We see that having a balanced training set with \(m=1\) is not the optimal solution for our approach. On the contrary, having an unbalanced training set improves its performance. From these observations, in the following experiments we train our models with a value of \(m\) fixed at 6.\footnote{It takes about 1.5 hours to train an epoch with the model settings described in~\ref{sssec:training-1} using a single GTX 1080 Ti GPU.}

\subsubsection{Evaluation on noisy parallel corpora}
\label{sssec:eval-comparaison}
\begin{figure}[t]
  \centering
  \includegraphics[width=0.45\textwidth]{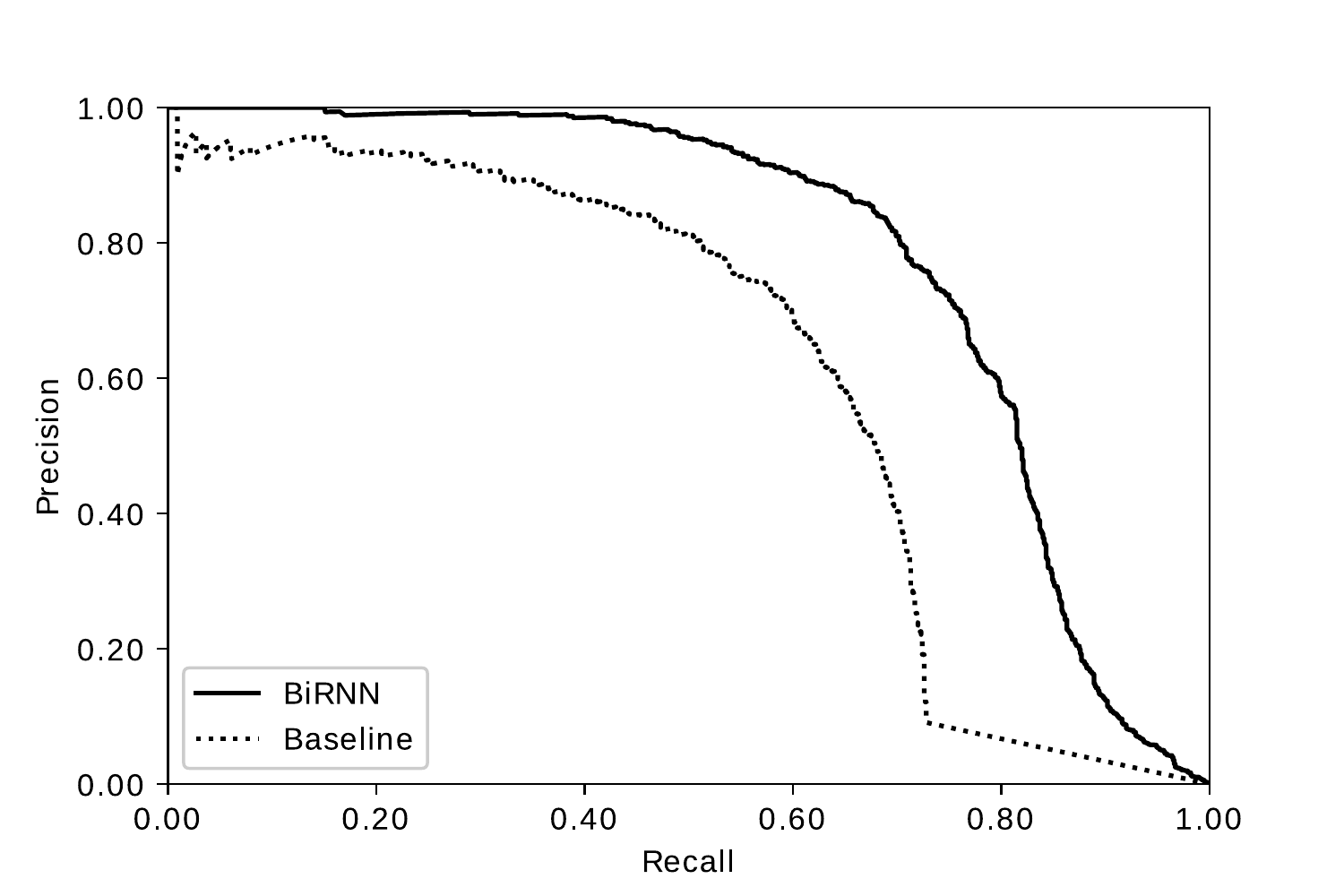}
  \includegraphics[width=0.45\textwidth]{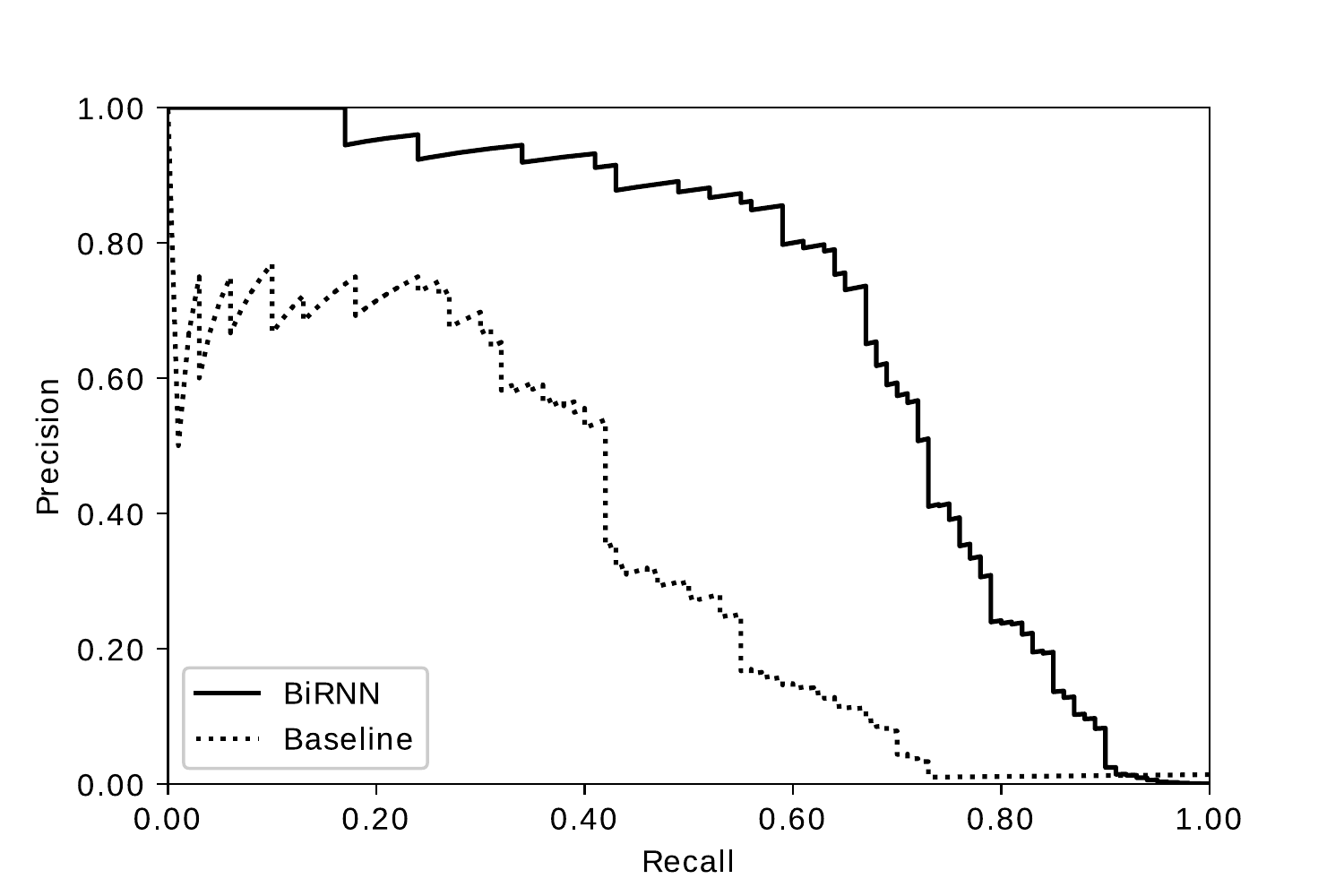}
  \caption{Precision-recall curves of the systems evaluated on newstest2012 with noise ratios of 0\% (left) and 90\% (right).}
  \label{fig:precision-recall}
\end{figure}

In this experiment, we compare our approach to the baseline system described in Section~\ref{sssec:baseline-1}. Table~\ref{table:comparaison} shows the precision, recall and F\(_{1}\) scores for the two systems evaluated on the newstest2012 and Europarl test sets with noise ratios of 0\%, 50\% and 90\%. We see that BiRNN is able to constantly outperform the results obtained with the baseline by a significant margin. By using newstest2012 as out-of-domain test sets, our approach gets an absolute improvement in the F\(_{1}\) score of 10.99\%, 13.03\% and 23.53\%. The precision-recall curves of the two systems evaluated on test sets with noise ratios of 0\% and 90\% are shown in Figure~\ref{fig:precision-recall}. These curves illustrate the consistency of our approach, whereas the performance of the baseline is greatly impacted when the number of non-parallel sentences in the test set is high.
\begin{table}[h!]
  \centering
  \fontsize{10}{12}\selectfont
  \begin{tabular}{*{1}{c}*{1}{l}*{4}{c}*{1}{c}*{4}{c}}
    \toprule 
    & & \multicolumn{4}{c}{\textbf{newstest2012}} & & \multicolumn{4}{c}{\textbf{Europarl}} \\
    \cmidrule(lr){3-6} \cmidrule(lr){8-11}
    \textbf{Noise}
      & \textbf{Model} 
      & \multicolumn{1}{c}{\textbf{P (\%)}} & \multicolumn{1}{c}{\textbf{R (\%)}}
      & \multicolumn{1}{c}{\textbf{F\(_{1}\) (\%)}} & \multicolumn{1}{c}{\textbf{\(\boldsymbol{\rho}\)}} 
      &
      & \multicolumn{1}{c}{\textbf{P (\%)}} & \multicolumn{1}{c}{\textbf{R (\%)}}
      & \multicolumn{1}{c}{\textbf{F\(_{1}\) (\%)}} & \multicolumn{1}{c}{\textbf{\(\boldsymbol{\rho}\)}} \\
    \midrule
    \addlinespace[5pt]
    \multirow{2}{*}{0\%} & BiRNN & \textbf{83.72} & \textbf{68.90} & \textbf{75.79} & 0.99 &
    						     & \textbf{99.26} & \textbf{93.50} & \textbf{96.29} & 0.99 \\
                         & Baseline & 73.88 & 57.70 & 64.80 & 0.93 &
          						    & 87.72 & 84.30 & 85.98 & 0.98 \\
    \midrule
    \multirow{2}{*}{50\%} & BiRNN & \textbf{79.95} & \textbf{65.40} & \textbf{71.95} & 0.99 &
                                 & \textbf{98.32} & \textbf{93.60} & \textbf{95.90} & 0.99 \\  
                         & Baseline & 73.43 & 49.20 & 58.92 & 0.98 &
							        & 80.19 & 82.60 & 81.38 & 0.99 \\
    \midrule
    \multirow{2}{*}{90\%} & BiRNN & \textbf{79.01} & \textbf{64.00} & \textbf{70.72} & 0.99 &
								 & \textbf{97.94} & \textbf{95.00} & \textbf{96.45} & 0.99 \\
                         & Baseline & 53.85 & 42.00 & 47.19 & 0.99 &
           						    & 63.89 & 69.00 & 66.35 & 0.99 \\                   
    \bottomrule
  \end{tabular}
  \caption{Precision (P), recall (R) and F\(_{1}\) scores where the decision threshold \(\rho\) maximizes the area under the precision-recall curve of the test sets with noise ratios of 0\%, 50\% and 90\%.}
  \label{table:comparaison}
\end{table}

\begin{figure}[h!]
  \centering
  \includegraphics[width=0.32\textwidth]{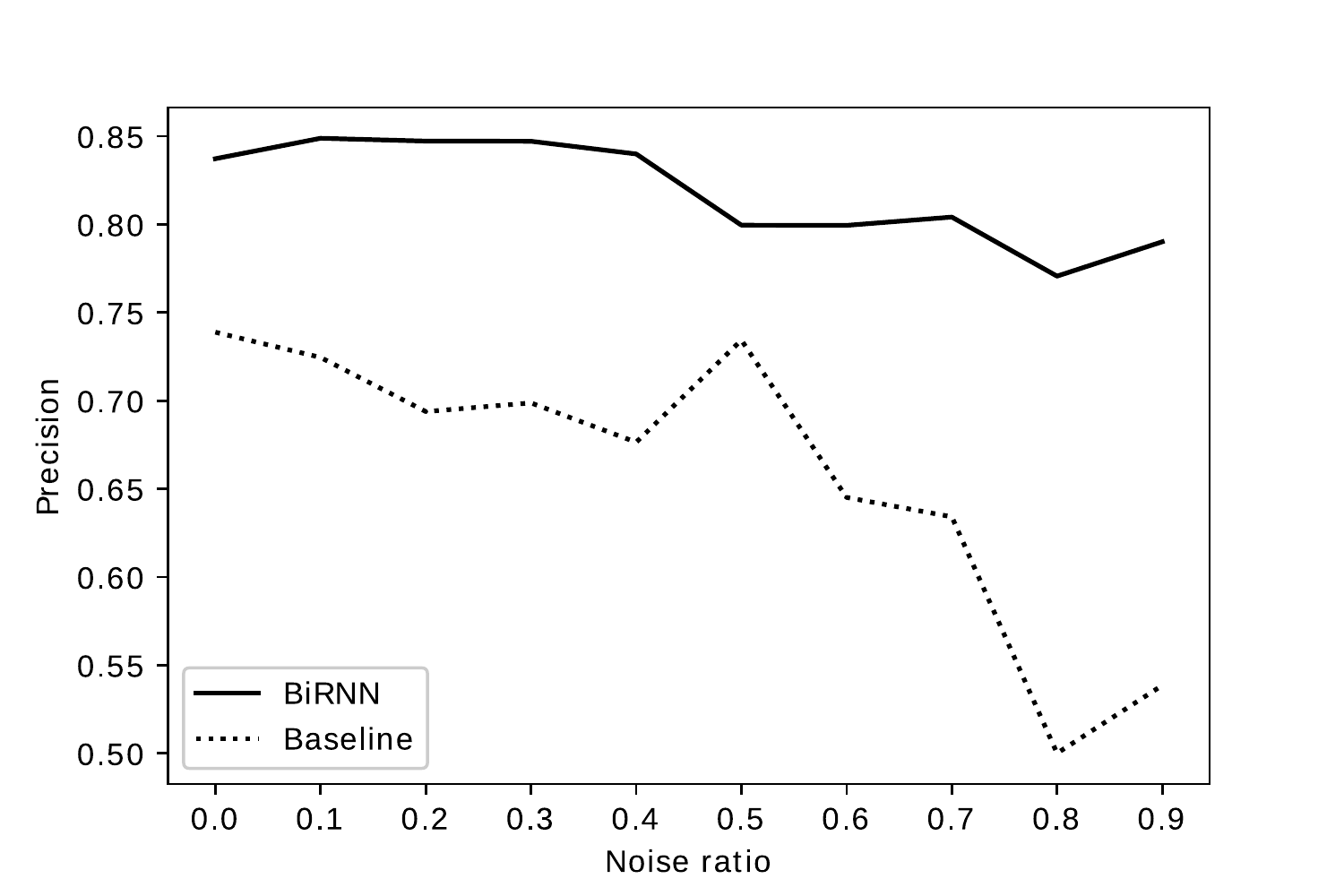}
  \includegraphics[width=0.32\textwidth]{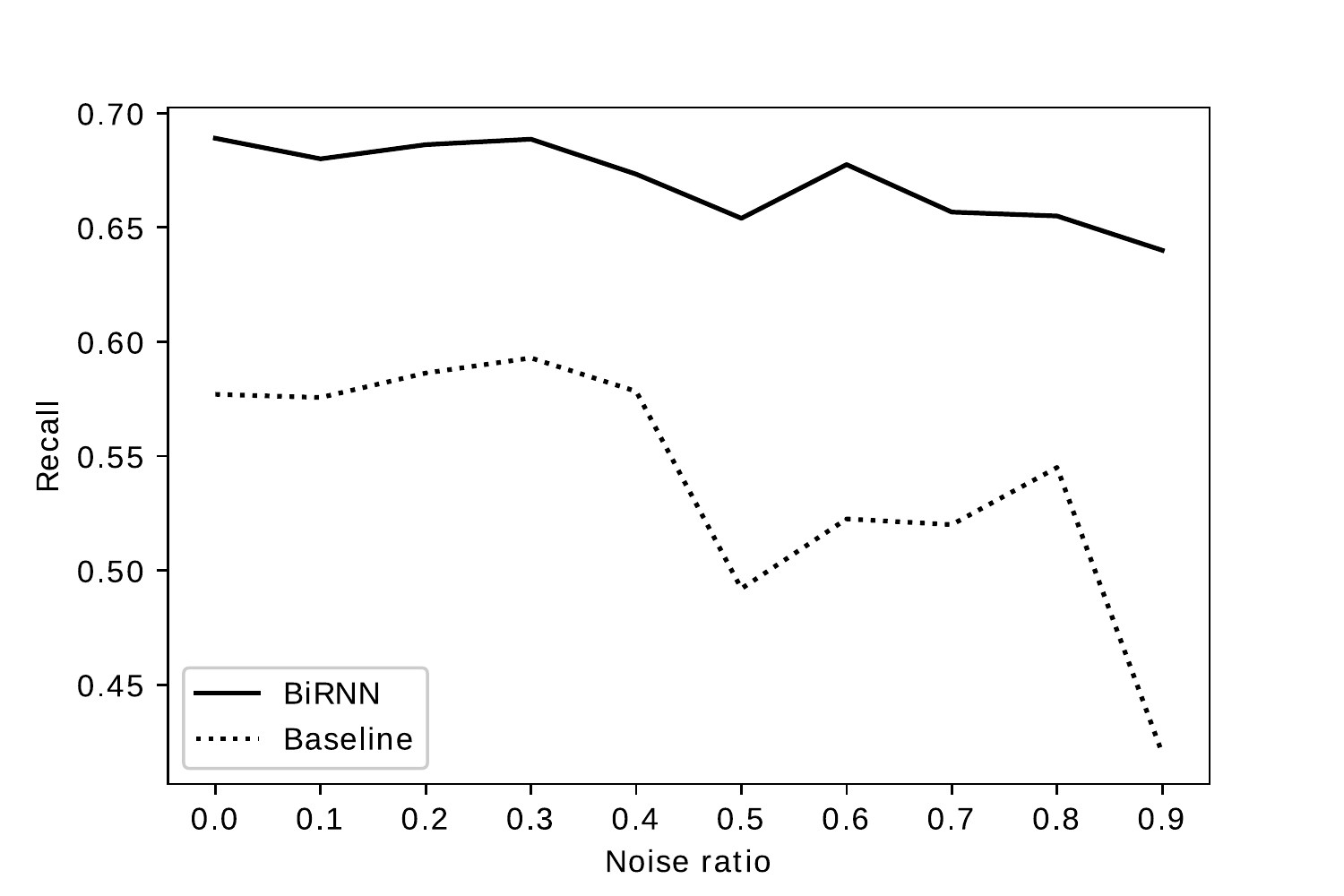}
  \includegraphics[width=0.32\textwidth]{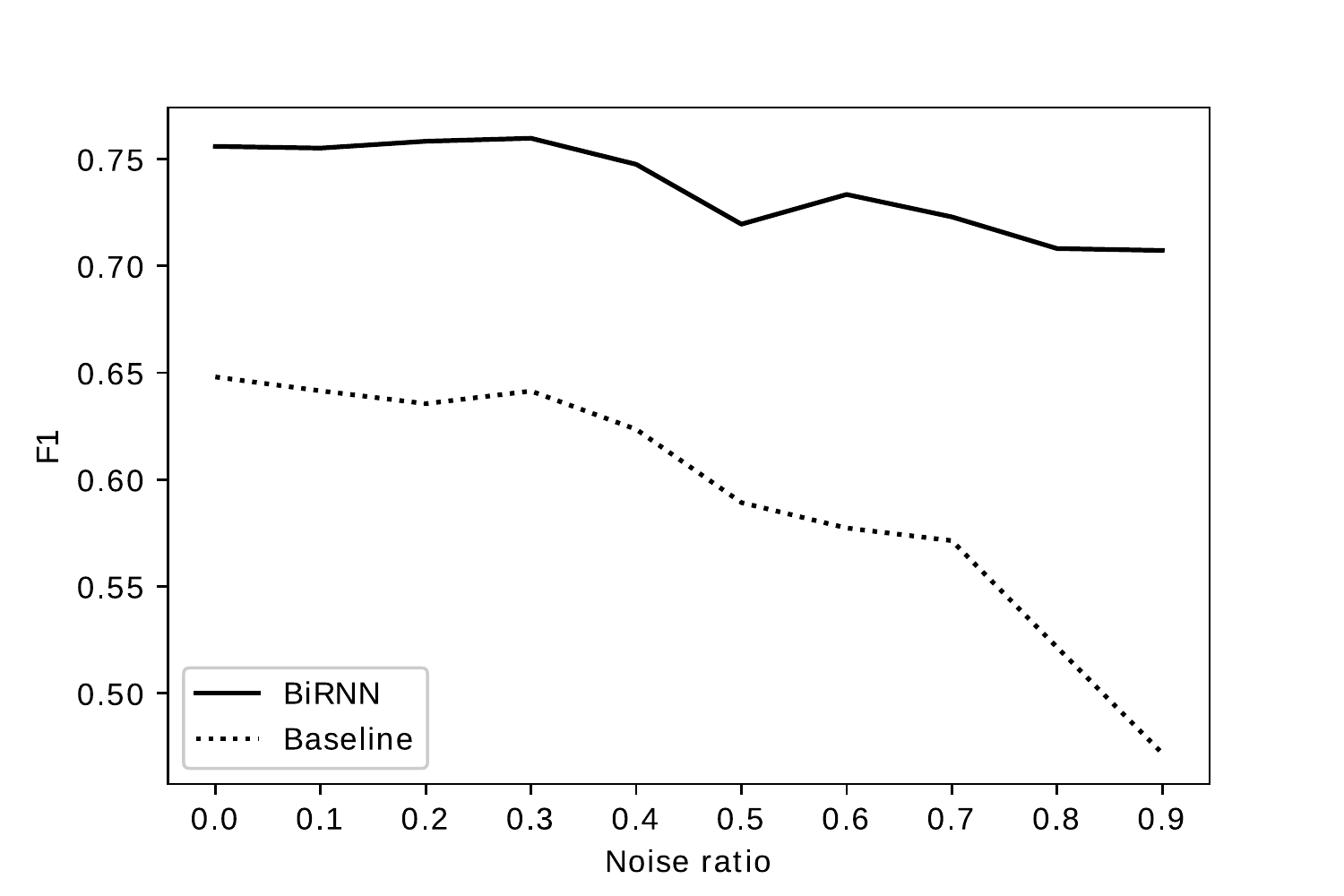}
  \caption{Precision (left), recall (middle) and F\(_{1}\) (right) scores of the systems evaluated on newstest2012 as the noise ratio \(r\) increases.}
  \label{fig:noise}
\end{figure}

When BiRNN is evaluated on Europarl, it obtains F\(_{1}\) scores over 95\% on the three test sets. On the other hand, given that the underlying models of the baseline system are trained on Europarl data, it is surprising to observe again a significant deterioration in its performance when the number of non-parallel sentences increases. In Figure~\ref{fig:noise}, we compare the precision, recall and F\(_{1}\) scores as the noise ratio \(r\) applied to our newstest2012 test set increases.  We observe that it becomes harder to identify parallel sentences as the number of non-parallel sentences increases in the test set. However, we see that our neural network based approach obtains better performances by a significant margin. In contrast to the baseline, the performance of our method stays relatively stable and starts to degrade at very high noise ratios. At that level of noise, we believe our test set is more representative to texts found in real comparable corpora. 

In this experiment, all our evaluation metrics (BiRNN and Baseline) are reported at the value \(\rho\) which maximizes the area under a precision-recall curve. These performances are therefore upper bound results. The decision threshold has a direct impact on the quality and the quantity of the extracted sentence pairs. Through our experiments, we observed that the optimal value \(\rho\) of BiRNN was constantly around a value of 0.99, while the one of the baseline system varied considerably from one test set to another. Thus, the stability of our approach is preferable in practice. We believe that the high optimal decision threshold value \(\rho = 0.99\) is caused by the fact that we use a sigmoid output layer with highly unbalanced training and test sets.

\subsection{Machine Translation Comparison}
\label{ssec:machine-translation}
\subsubsection{Data}
\label{sssec:data-2}
As training sets, we use the English--French Europarl corpus and sentence pairs extracted by the two parallel sentence extraction systems. The comparable corpora used to extract parallel sentences are 919,000 pairs of English--French articles obtained from Wikipedia dumps.\footnote{https://dumps.wikimedia.org/} The newstest2012 corpus is our validation set and we evaluate the machine translation systems on the newstest2013 corpus. All datasets are tokenized using the scripts from Moses. The maximum length of each sentence is set to 80 tokens. For the NMT systems, the vocabulary size for both languages is limited to the 150,000 most frequent words of their corresponding training set. All unknown words are replaced by the UNK token.

\subsubsection{Evaluation metric}
\label{sssec:eval-metrics-2}
To evaluate the translation performance of our systems we use the BLEU score~\cite{Papineni:2002} using the multi-bleu script from Moses.

\subsubsection{Training settings}
\label{sssec:training-2}
The SMT systems are phrase-based systems~\cite{Koehn:2003} that are trained with Moses. We use the GIZA++ implementation~\cite{Och:2003}\footnote{http://www.statmt.org/moses/giza/GIZA++.html} to train word alignment models in both directions. The phrases and lexical reordering are extracted using the default values of Moses. The language models are 5-gram models learned using the KenLM toolkit~\cite{KenLM:2011}\footnote{https://github.com/kpu/kenlm} on the monolingual parts of the same parallel corpus used for training the translation models. The parameters are optimized on the newstest2012 corpus. 

To train the NMT systems, we use the PyTorch implementation of OpenNMT~\cite{opennmt:2017}.\footnote{https://github.com/OpenNMT/OpenNMT-py} The NMT systems are one layer BiLSTMs with an attention mechanism~\cite{Bahdanau:2014}. The dimensions of the word embeddings and recurrent states for the encoder and decoder are all set to 256. The systems are trained for 10 epochs with minibatch of 64 sentence pairs using SGD with an initial learning rate of 1.0 and linear decay.\footnote{We could have trained NMT systems for more epochs to obtain better BLEU scores, but our objective is not to compare SMT with NMT systems.} The norm of the gradient is clipped such that it is not greater than 5.

\subsubsection{Evaluation on machine translation systems}
\label{sssec:eval-translation}
For each of the SMT and NMT approaches, we trained 14 machine translation systems. The first two systems,\footnote{That is, the two first systems for each approach and not from the total of 28 machine translation systems.} which serve as reference systems, are trained with 500,000 and 2,000,000 parallel sentences from the Europarl corpus, respectively. The system trained with 2,000,000 parallel sentences is used to compare the benefits of the sentence pairs extracted from Wikipedia against a large training corpus that we easily have at hand. For the 12 remaining systems, we sort the sentence pairs extracted by an extraction system in descending order according to the probability scores and append the top \(\{250000, 500000, \dots, 1250000, 1500000\}\) sentence pairs to the training set containing 500,000 parallel sentences from the Europarl corpus.

Table~\ref{table:bleu-scores} shows the BLEU scores obtained by the machine translation systems we trained for the SMT and NMT approaches. The figures in parentheses show the absolute gains in BLEU score compared to the reference systems trained with 500,000 parallel sentences from the Europarl corpus.\footnote{Which is the first line of Table~\ref{table:bleu-scores}.} We see that adding the subsets of sentence pairs extracted by both extraction systems leads to significant gains compared to the reference systems. Adding 1,500,000 sentence pairs extracted by BiRNN improves the BLEU score of the SMT and NMT systems by 3.71 and 9.47, respectively. We observe that translation systems trained on 1,000,000 sentence pairs have better BLEU scores than the reference systems trained on 2,000,000 parallel sentences from Europarl, showing that parallel sentence extraction allows translation systems to generalize better with less data. We are surprised that the BLEU scores obtained with the BiRNN data are so close to those obtained with the data from the baseline system. This may indicate that article pairs from Wikipedia share a relatively high degree of similarity. Still, the gain obtained with BiRNN compared to the baseline extraction system is consistent for all machine translation systems. These results confirm the quality of the 1,500,000 extracted sentence pairs. Hence, we could reduce the value of the decision threshold in order to extract corpora of larger sizes. Given the out-of-domain nature of the Wikipedia article pairs, we see that our approach could be applied to comparable corpora with a lower degree of comparability.
\begin{table}[t]
  \centering
  \fontsize{10}{12}\selectfont
  \setlength{\tabcolsep}{9.0pt}
  \begin{tabular}{*{2}{l}*{2}{c}*{1}{r}}
    \toprule 
    & & \multicolumn{2}{c}{\textbf{BLEU}} & \\
    \cmidrule(lr){3-4} 
    \multicolumn{1}{c}{\textbf{Data}} & \multicolumn{1}{c}{\textbf{Model}} 
    & \multicolumn{1}{c}{\textbf{SMT}} & \multicolumn{1}{c}{\textbf{NMT}} & \multicolumn{1}{c}{\textbf{Pairs}} \\
    \midrule
    \multirow{2}{*}{Europarl} & & 21.47 & 17.63 & 500k \\
                              & & 23.18 & 25.36 & 2M \\
    \midrule
    \multirow{2}{*}{+Top250k} & BiRNN & \textbf{23.11 (+1.64)} & \textbf{24.44 (+6.81)} & 750k \\
    					      & Baseline & 23.09 (+1.62) & 24.22 (+6.59) & 750k \\
    \midrule
    \multirow{2}{*}{+Top500k} & BiRNN & \textbf{24.12 (+2.65)} & \textbf{25.93 (+8.30)} & 1M \\
    					      & Baseline & 23.96 (+2.49) & 25.82 (+8.19) & 1M \\
    \midrule 
    \multirow{2}{*}{+Top750k} & BiRNN & \textbf{24.53 (+3.06)} & \textbf{26.39 (+8.76)} & 1.25M \\
    					      & Baseline & 24.44 (+2.97) & 26.19 (+8.56) & 1.25M \\
    \midrule 
    \multirow{2}{*}{+Top1M} & BiRNN & \textbf{24.85 (+3.38)} & \textbf{26.64 (+9.01)} & 1.5M \\
    					    & Baseline & 24.66 (+3.19) & 26.59 (+8.96) & 1.5M \\
    \midrule                       
    \multirow{2}{*}{+Top1.25M} & BiRNN & \textbf{25.01 (+3.54)} & \textbf{26.80 (+9.17)} & 1.75M \\
    					       & Baseline & 24.72 (+3.25) & 26.64 (+9.01) & 1.75M \\      
    \midrule                       
    \multirow{2}{*}{+Top1.5M} & BiRNN & \textbf{25.18 (+3.71)} & \textbf{27.10 (+9.47)} & 2M \\
    					      & Baseline & 25.01 (+3.54) & 26.85 (+9.22) & 2M \\                                 
    \bottomrule
  \end{tabular}
  \caption{BLEU scores obtained by the SMT and NMT systems on newstest2013. Pairs is the number of sentence pairs in a training set. The numbers in parentheses are the translation gains with respect to the reference system trained with 500,000 sentence pairs from the Europarl corpus.}
  \label{table:bleu-scores}
\end{table}

\section{Discussion}
\label{sect:discussion}
Most parallel sentence extraction systems developed so far were based on a series of models that required the computation of several features that did not necessarily adapt well to different text structures. To alleviate this problem, we have shown that we can create new parallel corpora from a collection of texts in different languages with a single end-to-end model based on bidirectional recurrent neural networks. Our approach deals directly with raw sentence pairs to determine if two sentences are a translation of each other. In addition to being more flexible than traditional systems, we demonstrate that our approach obtains better results than a competitive baseline system. We found that new sentence pairs extracted from comparable corpora is a considerable resource to exploit for SMT and NMT systems.

Our parallel sentence extraction system is completely based on neural network models and we believe that our work enables exploration for researchers who want to apply future research ideas. In this study we did not consider out-of-vocabulary words. Therefore, it would be beneficial to use word segments~\cite{Sennrich:2015} as lexical units instead of words to limit the size of the vocabulary and to be more robust when handling out-of-domain texts found in comparable corpora. The main challenge with our approach (and the majority of parallel sentence extraction systems) is that it needs to be trained on a parallel corpus. Thus, we need parallel sentences to extract parallel sentences, which can be problematic. Despite that it is possible to start from a small parallel corpus and to bootstrap the learning process with extracted sentence pairs, this method is limited for the many language pairs having few resources available. In order to lessen the need of a parallel corpus, an interesting and promising avenue could be to apply methods from some of the recent works~\cite{Xia:2016,Conneau:2017,Artetxe:2018} which allow machine translation systems to learn directly from non-parallel texts. Although our simple neural network architecture was able to extract parallel sentence pairs from comparable corpora, we believe that more sophisticated approaches, such as using two different sentence encoders trained by mutual information neural estimation (MINE)~\cite{Belghazi:2018} could improve the flexibility and the performance of our approach. Even though we believe that our approach is easily scalable across multiple language pairs, in this paper we only evaluated it on English--French texts. Therefore, it would be interesting to study a larger set of language pairs. In addition, we emphasize that it would be pertinent to evaluate our approach with distant language pairs which are poor in parallel resources. Our code is available on GitHub and the extracted parallel corpora will be released to the public.\footnote{https://github.com/FrancisGregoire/parSentExtract}

\section*{Acknowledgements}
This research was partially supported by the Natural Sciences and Engineering Research Council of Canada. 

\bibliographystyle{acl}
\bibliography{coling2018}

\begin{thebibliography}{}

\bibitem[\protect\citename{Abadi \bgroup et al.\egroup }2016]{Tensorflow:2016}
Mart\'{\i}n Abadi, Paul Barham, Jianmin Chen, Zhifeng Chen, Andy Davis, Jeffrey
  Dean, Matthieu Devin, Sanjay Ghemawat, Geoffrey Irving, Michael Isard,
  Manjunath Kudlur, Josh Levenberg, Rajat Monga, Sherry Moore, Derek~G. Murray,
  Benoit Steiner, Paul Tucker, Vijay Vasudevan, Pete Warden, Martin Wicke, Yuan
  Yu, and Xiaoqiang Zheng.
\newblock 2016.
\newblock Tensorflow: A system for large-scale machine learning.
\newblock In {\em Proceedings of the 12th USENIX Conference on Operating
  Systems Design and Implementation}, OSDI'16, pages 265--283, Berkeley, CA,
  USA. USENIX Association.

\bibitem[\protect\citename{Abdul-Rauf and Schwenk}2009]{Rauf:2009}
Sadaf Abdul-Rauf and Holger Schwenk.
\newblock 2009.
\newblock On the use of comparable corpora to improve smt performance.
\newblock In {\em Proceedings of the 12th Conference of the European Chapter of
  the Association for Computational Linguistics}, EACL '09, pages 16--23,
  Stroudsburg, PA, USA. Association for Computational Linguistics.

\bibitem[\protect\citename{Adafre and Rijke}2006]{Adafre:2006}
Sisay~Fissaha Adafre and Maarten~de Rijke.
\newblock 2006.
\newblock Finding similar sentences across multiple languages in wikipedia.
\newblock In {\em Proceedings of the Workshop on NEW TEXT Wikis and blogs and
  other dynamic text sources}.

\bibitem[\protect\citename{Artetxe \bgroup et al.\egroup }2017]{Artetxe:2018}
Mikel Artetxe, Gorka Labaka, Eneko Agirre, and Kyunghyun Cho.
\newblock 2017.
\newblock Unsupervised neural machine translation.
\newblock {\em CoRR}, abs/1710.11041.

\bibitem[\protect\citename{Azpeitia \bgroup et al.\egroup }2017]{Azpeitia:2017}
Andoni Azpeitia, Thierry Etchegoyhen, and Eva Mart{\'i}nez~Garcia.
\newblock 2017.
\newblock Weighted set-theoretic alignment of comparable sentences.
\newblock In {\em Proceedings of the 10th Workshop on Building and Using
  Comparable Corpora}, pages 41--45. Association for Computational Linguistics.

\bibitem[\protect\citename{Bahdanau \bgroup et al.\egroup }2014]{Bahdanau:2014}
Dzmitry Bahdanau, Kyunghyun Cho, and Yoshua Bengio.
\newblock 2014.
\newblock Neural machine translation by jointly learning to align and
  translate.
\newblock {\em CoRR}, abs/1409.0473.

\bibitem[\protect\citename{Belghazi \bgroup et al.\egroup }2018]{Belghazi:2018}
Ishmael Belghazi, Sai Rajeswar, Aristide Baratin, R.~Devon Hjelm, and Aaron~C.
  Courville.
\newblock 2018.
\newblock {MINE:} mutual information neural estimation.
\newblock {\em CoRR}, abs/1801.04062.

\bibitem[\protect\citename{Bromley \bgroup et al.\egroup }1993]{Bromley:1993}
Jane Bromley, Isabelle Guyon, Yann LeCun, Eduard S\"{a}ckinger, and Roopak
  Shah.
\newblock 1993.
\newblock Signature verification using a "siamese" time delay neural network.
\newblock In {\em Proceedings of the 6th International Conference on Neural
  Information Processing Systems}, NIPS'93, pages 737--744, San Francisco, CA,
  USA. Morgan Kaufmann Publishers Inc.

\bibitem[\protect\citename{Brown \bgroup et al.\egroup }1993]{Brown:1993}
Peter~F. Brown, Vincent J.~Della Pietra, Stephen A.~Della Pietra, and Robert~L.
  Mercer.
\newblock 1993.
\newblock The mathematics of statistical machine translation: Parameter
  estimation.
\newblock {\em Comput. Linguist.}, 19(2):263--311, June.

\bibitem[\protect\citename{Cho \bgroup et al.\egroup }2014]{Cho:2014}
Kyunghyun Cho, Bart van Merrienboer, {\c{C}}aglar G{\"{u}}l{\c{c}}ehre, Fethi
  Bougares, Holger Schwenk, and Yoshua Bengio.
\newblock 2014.
\newblock Learning phrase representations using rnn encoder-decoder for
  statistical machine translation.
\newblock {\em CoRR}, abs/1406.1078.

\bibitem[\protect\citename{Chu \bgroup et al.\egroup }2016]{Chu:2016}
Chenhui Chu, Raj Dabre, and Sadao Kurohashi.
\newblock 2016.
\newblock Parallel sentence extraction from comparable corpora with neural
  network features.
\newblock In Nicoletta Calzolari~(Conference Chair), Khalid Choukri, Thierry
  Declerck, Sara Goggi, Marko Grobelnik, Bente Maegaard, Joseph Mariani, Helene
  Mazo, Asuncion Moreno, Jan Odijk, and Stelios Piperidis, editors, {\em
  Proceedings of the Tenth International Conference on Language Resources and
  Evaluation (LREC 2016)}, Paris, France, may. European Language Resources
  Association (ELRA).

\bibitem[\protect\citename{Conneau \bgroup et al.\egroup }2017]{Conneau:2017}
Alexis Conneau, Guillaume Lample, Marc'Aurelio Ranzato, Ludovic Denoyer, and
  Herv{\'{e}} J{\'{e}}gou.
\newblock 2017.
\newblock Word translation without parallel data.
\newblock {\em CoRR}, abs/1710.04087.

\bibitem[\protect\citename{Espa{\~{n}}a{-}Bonet \bgroup et al.\egroup
  }2017]{Cristina:2017}
Cristina Espa{\~{n}}a{-}Bonet, {\'{A}}d{\'{a}}m~Csaba Varga, Alberto
  Barr{\'{o}}n{-}Cede{\~{n}}o, and Josef van Genabith.
\newblock 2017.
\newblock An empirical analysis of nmt-derived interlingual embeddings and
  their use in parallel sentence identification.
\newblock {\em CoRR}, abs/1704.05415.

\bibitem[\protect\citename{Gouws \bgroup et al.\egroup }2015]{Gouws:2015}
Stephan Gouws, Yoshua Bengio, and Greg Corrado.
\newblock 2015.
\newblock Bilbowa: Fast bilingual distributed representations without word
  alignments.
\newblock In {\em Proceedings of the 32Nd International Conference on
  International Conference on Machine Learning - Volume 37}, ICML'15, pages
  748--756. JMLR.org.

\bibitem[\protect\citename{Graves}2013]{Graves:2013}
Alex Graves.
\newblock 2013.
\newblock Generating sequences with recurrent neural networks.
\newblock {\em CoRR}, abs/1308.0850.

\bibitem[\protect\citename{Grover and Mitra}2017]{Grover:2017}
Jeenu Grover and Pabitra Mitra.
\newblock 2017.
\newblock Bilingual word embeddings with bucketed cnn for parallel sentence
  extraction.
\newblock In {\em Proceedings of ACL 2017, Student Research Workshop}, pages
  11--16. Association for Computational Linguistics.

\bibitem[\protect\citename{Heafield}2011]{KenLM:2011}
Kenneth Heafield.
\newblock 2011.
\newblock Kenlm: Faster and smaller language model queries.
\newblock In {\em Proceedings of the Sixth Workshop on Statistical Machine
  Translation}, pages 187--197. Association for Computational Linguistics.

\bibitem[\protect\citename{Hermann \bgroup et al.\egroup }2015]{Hermann:2015}
Karl~Moritz Hermann, Tom{\'{a}}s Kocisk{\'{y}}, Edward Grefenstette, Lasse
  Espeholt, Will Kay, Mustafa Suleyman, and Phil Blunsom.
\newblock 2015.
\newblock Teaching machines to read and comprehend.
\newblock {\em CoRR}, abs/1506.03340.

\bibitem[\protect\citename{Hochreiter and Schmidhuber}1997]{Hochreiter:97}
Sepp Hochreiter and J\"{u}rgen Schmidhuber.
\newblock 1997.
\newblock Long short-term memory.
\newblock {\em Neural Compututation}, 9(8):1735--1780, November.

\bibitem[\protect\citename{Kingma and Ba}2014]{Kingma:2014}
Diederik~P. Kingma and Jimmy Ba.
\newblock 2014.
\newblock Adam: {A} method for stochastic optimization.
\newblock {\em CoRR}, abs/1412.6980.

\bibitem[\protect\citename{Klein \bgroup et al.\egroup }2017]{opennmt:2017}
Guillaume Klein, Yoon Kim, Yuntian Deng, Jean Senellart, and Alexander~M. Rush.
\newblock 2017.
\newblock Opennmt: Open-source toolkit for neural machine translation.
\newblock {\em CoRR}, abs/1701.02810.

\bibitem[\protect\citename{Koehn \bgroup et al.\egroup }2003]{Koehn:2003}
Philipp Koehn, Franz~Josef Och, and Daniel Marcu.
\newblock 2003.
\newblock Statistical phrase-based translation.
\newblock In {\em Proceedings of the 2003 Conference of the North American
  Chapter of the Association for Computational Linguistics on Human Language
  Technology - Volume 1}, NAACL '03, pages 48--54, Stroudsburg, PA, USA.
  Association for Computational Linguistics.

\bibitem[\protect\citename{Koehn \bgroup et al.\egroup }2007]{Moses:2007}
Philipp Koehn, Hieu Hoang, Alexandra Birch, Chris Callison-Burch, Marcello
  Federico, Nicola Bertoldi, Brooke Cowan, Wade Shen, Christine Moran, Richard
  Zens, Chris Dyer, Ond\v{r}ej Bojar, Alexandra Constantin, and Evan Herbst.
\newblock 2007.
\newblock Moses: Open source toolkit for statistical machine translation.
\newblock In {\em Proceedings of the 45th Annual Meeting of the ACL on
  Interactive Poster and Demonstration Sessions}, ACL '07, pages 177--180,
  Stroudsburg, PA, USA. Association for Computational Linguistics.

\bibitem[\protect\citename{Koehn}2005]{Koehn:2005}
Philipp Koehn.
\newblock 2005.
\newblock {Europarl: A Parallel Corpus for Statistical Machine Translation}.
\newblock In {\em {Conference Proceedings: the tenth Machine Translation
  Summit}}, pages 79--86, Phuket, Thailand. AAMT, AAMT.

\bibitem[\protect\citename{Luong \bgroup et al.\egroup }2015]{Luong:2015}
Minh-Thang Luong, Hieu Pham, and Christopher~D. Manning.
\newblock 2015.
\newblock Bilingual word representations with monolingual quality in mind.
\newblock In {\em NAACL Workshop on Vector Space Modeling for NLP}, Denver,
  United States.

\bibitem[\protect\citename{Munteanu and Marcu}2005]{Munteanu:2005}
Dragos~Stefan Munteanu and Daniel Marcu.
\newblock 2005.
\newblock Improving machine translation performance by exploiting non-parallel
  corpora.
\newblock {\em Computational Linguistics}, 31(4):477--504, December.

\bibitem[\protect\citename{Och and Ney}2003]{Och:2003}
Franz~Josef Och and Hermann Ney.
\newblock 2003.
\newblock A systematic comparison of various statistical alignment models.
\newblock {\em Comput. Linguist.}, 29(1):19--51, March.

\bibitem[\protect\citename{Papineni \bgroup et al.\egroup }2002]{Papineni:2002}
Kishore Papineni, Salim Roukos, Todd Ward, and Wei-Jing Zhu.
\newblock 2002.
\newblock Bleu: A method for automatic evaluation of machine translation.
\newblock In {\em Proceedings of the 40th Annual Meeting on Association for
  Computational Linguistics}, ACL '02, pages 311--318, Stroudsburg, PA, USA.
  Association for Computational Linguistics.

\bibitem[\protect\citename{Pascanu \bgroup et al.\egroup }2013]{Pascanu:2013}
Razvan Pascanu, Tomas Mikolov, and Yoshua Bengio.
\newblock 2013.
\newblock On the difficulty of training recurrent neural networks.
\newblock In {\em Proceedings of the 30th International Conference on
  International Conference on Machine Learning - Volume 28}, ICML'13, pages
  III--1310--III--1318. JMLR.org.

\bibitem[\protect\citename{Schuster and Paliwal}1997]{Schuster:97}
Mike. Schuster and Kuldip~K. Paliwal.
\newblock 1997.
\newblock Bidirectional recurrent neural networks.
\newblock {\em Trans. Sig. Proc.}, 45(11):2673--2681.

\bibitem[\protect\citename{Sennrich \bgroup et al.\egroup }2015]{Sennrich:2015}
Rico Sennrich, Barry Haddow, and Alexandra Birch.
\newblock 2015.
\newblock Neural machine translation of rare words with subword units.
\newblock {\em CoRR}, abs/1508.07909.

\bibitem[\protect\citename{Smith \bgroup et al.\egroup }2010]{Smith:2010}
Jason~R. Smith, Chris Quirk, and Kristina Toutanova.
\newblock 2010.
\newblock Extracting parallel sentences from comparable corpora using document
  level alignment.
\newblock In {\em Human Language Technologies: The 2010 Annual Conference of
  the North American Chapter of the Association for Computational Linguistics},
  HLT '10, pages 403--411, Stroudsburg, PA, USA. Association for Computational
  Linguistics.

\bibitem[\protect\citename{Smith \bgroup et al.\egroup }2017]{Smith:2017}
Samuel~L. Smith, David H.~P. Turban, Steven Hamblin, and Nils~Y. Hammerla.
\newblock 2017.
\newblock Offline bilingual word vectors, orthogonal transformations and the
  inverted softmax.
\newblock {\em CoRR}, abs/1702.03859.

\bibitem[\protect\citename{Sutskever \bgroup et al.\egroup
  }2014]{Sutskever:2014}
Ilya Sutskever, Oriol Vinyals, and Quoc~V. Le.
\newblock 2014.
\newblock Sequence to sequence learning with neural networks.
\newblock In {\em Proceedings of the 27th International Conference on Neural
  Information Processing Systems}, NIPS'14, pages 3104--3112, Cambridge, MA,
  USA. MIT Press.

\bibitem[\protect\citename{Vogel \bgroup et al.\egroup }1996]{Vogel:1996}
Stephan Vogel, Hermann Ney, and Christoph Tillmann.
\newblock 1996.
\newblock Hmm-based word alignment in statistical translation.
\newblock In {\em Proceedings of the 16th Conference on Computational
  Linguistics - Volume 2}, COLING '96, pages 836--841, Stroudsburg, PA, USA.
  Association for Computational Linguistics.

\bibitem[\protect\citename{Xia \bgroup et al.\egroup }2016]{Xia:2016}
Yingce Xia, Di~He, Tao Qin, Liwei Wang, Nenghai Yu, Tie{-}Yan Liu, and
  Wei{-}Ying Ma.
\newblock 2016.
\newblock Dual learning for machine translation.
\newblock {\em CoRR}, abs/1611.00179.

\bibitem[\protect\citename{Zaremba \bgroup et al.\egroup }2014]{Zaremba:2014}
Wojciech Zaremba, Ilya Sutskever, and Oriol Vinyals.
\newblock 2014.
\newblock Recurrent neural network regularization.
\newblock {\em CoRR}, abs/1409.2329.

\end{thebibliography}

\end{document}